\documentclass[sigconf]{acmart}
\AtBeginDocument{}

\usepackage{enumitem}
\usepackage{wrapfig}
\usepackage{graphicx}
\usepackage{multirow}
\usepackage{adjustbox}
\usepackage[utf8]{inputenc}
\usepackage[T1]{fontenc}
\usepackage{hyperref}
\usepackage{url}
\usepackage{booktabs}
\usepackage{amsfonts}
\usepackage{nicefrac}
\usepackage{microtype}
\usepackage{xcolor}
\usepackage{amsmath}

\copyrightyear{2026}
\acmYear{2026}
\setcopyright{cc}
\setcctype{by}
\acmConference[SIGSPATIAL '26]{The 34th ACM International Conference on Advances in Geographic Information Systems}{November 03--06, 2026}{Riverside, CA, USA}
\acmBooktitle{The 34th ACM International Conference on Advances in Geographic Information Systems (SIGSPATIAL '26), November 03--06, 2026, Riverside, CA, USA}
\acmDOI{10.1145/3841645.3843371}
\acmISBN{979-8-4007-2950-8/2026/11}

\begin{document}

\title{Earth System World Model for What-If Simulations: A Case Study for Terrestrial Ecosystems}

\author{Zhihao Wang$^{1}$, Ruichen Wang$^{1}$, Ruohan Li$^{1}$, Lei Ma$^{1}$, George Hurtt$^{1}$, \\
Xiaowei Jia$^{2}$, Gengchen Mai$^{3}$, Shaowen Wang$^{4}$, Yiqun Xie$^{1}$}
\affiliation{
    $^1$University of Maryland,
    $^2$Rutgers University,
    $^3$University of Texas at Austin,
    $^4$University of Illinois Urbana-Champaign
    \country{}
}
\email{\{zhwang1, ruichenw, r526li, lma6, gchurtt, xie\}@umd.edu,\\
xj159@cs.rutgers.edu,
gengchen.mai@austin.utexas.edu,
shaowen@illinois.edu}
\authornote{Corresponding author: Yiqun Xie.\vspace{-5.1pt}}

\renewcommand{\shortauthors}{Wang et al.}

\begin{abstract}

Machine learning emulators have become essential for accelerating expensive Earth-system simulations, but most existing approaches remain passive forecasters: they reproduce simulator trajectories under prescribed forcings without an explicit interaction mechanism for user-specified interventions.
This limits their use in interactive scientific workflows and Earth-system digital twins, where users often need to explore how a system would respond if selected state components were changed.
We propose an action-conditioned world-modeling framework for Earth-system emulation that reformulates simulator trajectories as supervision for controllable state-transition learning.
The key idea is transition-action pretraining: naturally observed state changes are treated as label-free action supervision, allowing the model to learn both prescribed dynamics and action-conditioned responses without manually annotated interventions.
We further introduce masked response learning to infer unobserved variables under partial state edits and learn coupled system dependencies.
We test this framework on ecosystem dynamics across six global regions and multiple stand ages.
Experiments show that the model preserves competitive long-horizon emulation accuracy while enabling controllable structural interventions and coherent responses in coupled ecosystem-cycle variables.
These results suggest a practical route from passive Earth-system emulators
toward interactive, intervention-aware scientific surrogates.

\end{abstract}



\maketitle

\section{Introduction}

Machine learning emulators have become an important tool for accelerating expensive Earth-system models, including
atmospheric, terrestrial, and hydrological
simulations \cite{lam2023learning, wang2023high}. By learning surrogate mappings from environmental forcings and initial states to future system trajectories, these models can reduce the computational cost of large-scale simulations and enable scientific discovery under a much broader range of scenarios for impact assessment (i.e., answering ``what-if'' questions).
However, most Earth-system emulators are designed as passive forecasters or approximators: they reproduce model outputs under prescribed forcings, but they do not provide an explicit model design to enable user-specified interactions or interventions (e.g., disturbance events such as wildfire, logging, mortality, management actions, or policy-induced changes \cite{seidl2011unraveling, wang2025treefinder}).
This is analogous to early-stage video generators that can create high-fidelity pixel streams but do not provide interactability \cite{assran2025v}.
This limits the use of the machine learning emulators in interactive scientific and application workflows for Earth-system digital twins, where users are often interested in asking not only what trajectory is likely under a given scenario, but also how the system would respond if selected components of the state were changed \cite{morecroft2019measuring}.
Such \textbf{what-if reasoning} is
essential components of exploratory analysis and hypothesis generation in scientific studies and policy making,
yet less considered by standard emulators.

\begin{figure*}[ht!]
    \centering
    \includegraphics[width=1\linewidth]{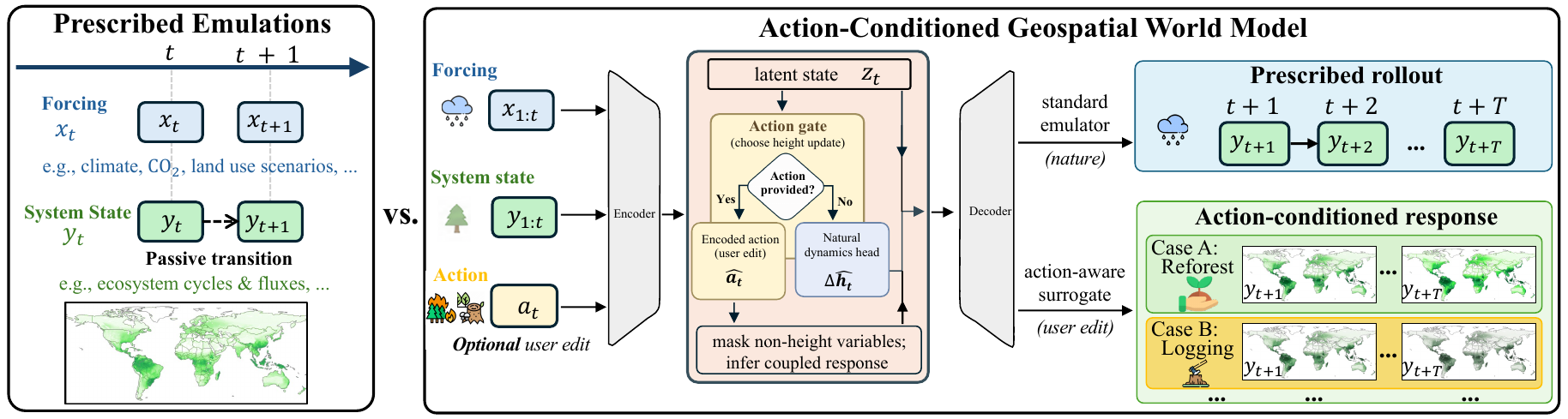}
    \vspace{-20pt}
    \caption{Overview of the proposed action-conditioned world-modeling framework for extending passive Earth-system emulators with controllable state edits, transition-action pretraining, and masked response inference.}
    \label{fig:overall}
    \vspace{-10pt}
\end{figure*}

Recent progress in world models provides a promising direction for moving beyond passive emulation.
World models aim to learn compact latent representations of system states and predict how the latent states evolve under actions, with interactability and controllability as their key distinctions: rather than only extrapolating from fixed inputs, the model should update its future predictions in response to interventions, actions, or goals injected on the fly \cite{assran2025v, ding2025understanding}.
This action-conditioned dynamics formulation is particularly relevant for Earth-system emulation, where a learned surrogate could support rapid exploration of alternative management and policy decisions, or structural latent state edits \cite{morecroft2019measuring}.
However, directly adapting existing world-model methods to Earth systems requires additional considerations and tailored designs.
Unlike games or robotics, Earth-system trajectories often lack explicit action labels, and many relevant interventions are not observed as discrete commands. Moreover, Earth-system states are multi-variable, partially observed, and physically coupled across variables and time scales, so an intervention on one component of the system should induce coherent responses in other variables.

In this study, we take an exploratory step toward \textbf{interactive world models} for Earth-system emulation, via a concrete case study on terrestrial ecosystems.
Specifically, we reformulate emulator training from passive input-output supervision into action-conditioned state-transition modeling, enabling rapid what-if exploration for explicit, user-edited system states and their coupled responses.
Our contributions are:
(1) We introduce transition-action pretraining, which derives label-free action supervision from naturally observed state changes;
(2) We integrate masked response learning as a required step to improve the model's stability  with partially unobserved variables, which is needed to enable partial state edits;
(3) We conduct experiments on representative ecosystem trajectories over heterogeneous regions from 6 continents.

\section{Related Work}

\textbf{Earth-system emulation.} Deep learning emulators have been increasingly developed to approximate expensive theory-based models across atmospheric, terrestrial, and hydrological processes \cite{lam2023learning, wang2023high, li2024solarcube}.
For example, in weather forecasting, AI emulators have shown strong potential for accelerating medium-range global prediction by learning directly from large-scale atmospheric reanalysis data such as ERA5 \cite{kurth2023fourcastnet}.
Similarly, for terrestrial ecosystems,
recent work has demonstrated orders of magnitude speed-ups from AI emulators for calibrated, high-fidelity large-scale simulations \cite{wang2025carbonglobe, li2026ecodiffusion}.
However, these emulators usually do not expose an explicit model interaction mechanism
for user-defined interventions or partial state edits.
As a result, they are less suited for interactive what-if analysis, where users need to modify selected system components dynamically on the fly and infer the coupled downstream response.

\textbf{World models and controllable dynamics.} World models provide a complementary direction by learning compact representations of system state and modeling how those representations evolve under actions \cite{ha2018world}.
They have been widely studied in reinforcement learning, robotics, video prediction, autonomous driving, and interactive generative environments, where action-conditioned dynamics enable on-the-fly planning, control, and counterfactual interaction \cite{huang2026pointworld}.
Recent latent-prediction and generative world models further show that future states can be predicted in representation space or generated as interactive environments, making controllability a central distinction from passive forecasting \cite{assran2025v, ding2025understanding}.
However,
unlike robotics or games, Earth-system trajectories typically lack explicit action labels, where the interventions are often continuous or implicit rather than directly usable discrete commands (e.g., turn left or right in video simulation), and system states are physically coupled
and partially observed

To the best of our knowledge, this work is among the first attempts to explore and bridge these two lines of research by reformulating AI-driven
Earth-system emulation as an action-conditioned, interactive world model.
Our approach uses transition-action pretraining and masked response learning to move beyond passive surrogate prediction toward interactive, intervention-aware scientific emulation.

\section{Method}

\subsection{Problem Formulation}

We consider Earth-system trajectories consisting of external forcings and system states.
Let $\mathbf{x}_t\in \mathbb{R}^{M}$ denote the environmental forcings (e.g., temperature, precipitation, CO$_2$) at time $t$, and $\mathbf{y}_t \in \mathbb{R}^{D}$ denote a $D$-dimensional system state.
Given an example of terrestrial ecosystems,
$\mathbf{y}_t$ contains coupled structural and ecosystem variables, including canopy height, above-ground biomass, soil carbon, ecosystem carbon fluxes, etc.

Standard emulators learn a passive transition model,
\begin{equation}
    \widehat{\mathbf{y}}_{t+1}
    =
    F_{\theta}\!\left(
        \mathbf{y}_{1:t},
        \mathbf{x}_{1:t}
    \right),
\end{equation}
which predicts future simulator outputs under prescribed forcings and historical observations.
In contrast, our goal is to learn an action-conditioned state-transition model,
\begin{equation}
    \widehat{\mathbf{y}}_{t+1}
    =
    F_{\theta}\!\left(
        \mathbf{y}_{1:t},
        \mathbf{x}_{1:t},
        \mathbf{a}_{t}
    \right),
\end{equation}
where $\mathbf{a}_{t}$ denotes a user-specified state-edit action that may be absent during prescribed rollout.
In Earth-system emulation, actions need not correspond to discrete commands as in games or robotics.
They can represent interventions that modify selected components of the system state, such as logging-induced reductions in vegetation structure, reforestation-driven structural recovery, disturbance events, or policy-induced land-management changes.

\subsection{Transition-Action Pretraining}

A central challenge in Earth-system emulation is that simulator trajectories usually do not contain explicit action labels.
We therefore derive action supervision directly from observed state transitions. Let $S \subseteq \{1,\ldots,D\}$ denote the subset of controlled variables in system states. For each transition $\mathbf{y}_{t} \rightarrow \mathbf{y}_{t+1}$, we define a transition-derived action as:
\begin{equation}\label{eq:action}
    \mathbf{a}^{S}_{t}
    =
    \mathbf{y}^{S}_{t+1}
    -
    \mathbf{y}^{S}_{t}.
\end{equation}
\noindent where $\mathbf{a}^{S}_{t}$, $\mathbf{y}^{S}_{t+1}$, and $\mathbf{y}^{S}_{t}$ denote the subset of controlled variables in $\mathbf{a}_{t}$, $\mathbf{y}_{t+1}$, and $\mathbf{y}_{t}$.
This converts every simulator transition into a self-supervised action example without requiring manually annotated interventions.
In this study, we instantiate the controlled variable as canopy height.
Let $h_t$ denote the canopy-height component of the system state $\mathbf{y}_t$. The transition-derived action is then defined as the canopy-height changes: $h_{t+1} - h_{t}$.

The action here provides a simple but physically meaningful structural intervention.
Positive actions correspond to accelerated structural growth, while negative actions can represent structural reductions such as harvesting or disturbance.
The remaining ecosystem variables are not directly interacted; instead, the model must infer their coupled response from the action-conditioned transition.

During training, the model alternates between two modes. In the prescribed mode,
it learns natural system evolution from forcings and state history without an action.
In the action-conditioned mode, a transition-derived action is provided, and the model learns to propagate the edited controlled state into the remaining variables.
This strategy allows the same simulator trajectories to supervise both passive emulation and controllable state-transition learning.

\subsection{Action-Conditioned Dynamics}

A key design challenge is to make the model controllable without losing its ability to predict natural system evolution.
If the action is represented only as a latent token, the network may bypass it because the same transition can often be inferred from state history and environmental forcings.
In contrast, if the controlled state is always defined directly by the action, the model becomes controllable but cannot predict that state when no action is provided.
We address this tension with a gated dynamics design that combines prescribed growth prediction with action-conditioned state editing.

Given the forcing history and observed state history, an encoder produces a latent representation $\mathbf{z}_{t}$.
A natural-dynamics head predicts the prescribed increment of the controlled variable:
\begin{equation}
    \Delta \widehat{h}_{t}
    =
    f_{\theta}\!\left(\mathbf{z}_{t}\right).
\end{equation}
The next controlled state is then computed and decoded by an action gate:
\begin{equation}
    \widehat{h}_{t+1}
    =
    \begin{cases}
    h_{t} + D_h (\Delta \widehat{h}_{t}), & \text{if no action is provided}, \\
    h_{t} + D_h (\widehat{a}_{t}), & \text{if provided ($\widehat{a}_t$ is the encoded action)}.
    \end{cases}
\end{equation}
The remaining variables are decoded conditioned on the latent representation and the resulting controlled state:
\begin{equation}
    \widehat{\mathbf{y}}^{-h}_{t+1}
    =
    g_{\theta}\!\left(
        \mathbf{z}_{t},
        \widehat{h}_{t+1}
    \right),
\end{equation}
where $\widehat{\mathbf{y}}^{-h}_{t+1}$ denotes all predicted variables except canopy height. The final predicted state is
\begin{equation}
    \widehat{\mathbf{y}}_{t+1}
    =
    \left[
        \widehat{h}_{t+1};
        \widehat{\mathbf{y}}^{-h}_{t+1}
    \right].
\end{equation}
This design gives a single model two operating modes: prescribed rollout when no action is provided and controllable response prediction when an action is specified.

\subsection{Masked Response Learning}

To infer the coupled response when an intervention modifies only part of the system state, we integrate masked response learning during action-conditioned training, which is necessary to improve the model's stability to handle user interventions that will change one or a subset of variables but leave others outdated.
Let $\bar{S}$ denote the complement subset (i.e., non-controlled) of the controlled variable subset $S$. In the action-conditioned mode, the model observes the controlled state and action while masking the non-controlled variables at the edited step, i.e., setting non-controlled variables in $\bar{S}$ at the input step $t$ to 0.
This reflects the scenarios that will be encountered during the inference, where users may edit controlled variables and the model will need to predict with the outdated non-controlled variables masked out.
Additionally, in the action-conditioned training mode (randomly assigned to samples), the "true reference" of controlled variables in $S$ at the next step is already provided by the actions simulated using Eq. (\ref{eq:action}), so the loss function will have these variables masked out in loss calculation.
During inference, the model can be rolled out without actions as a standard emulator or run with interactive user state edits to examine controlled responses.

\vspace{-8pt}
\section{Experiment}

\vspace{-2pt}
\subsection{Experimental Setup}

\textbf{Dataset.}
We evaluate our framework on CarbonGlobe \cite{wang2025carbonglobe}, a global, 40-years ecosystem forecasting dataset based on the Ecosystem Demography (ED) model.
We select six globally distributed regions covering heterogeneous ecosystem conditions and evaluate three representative stand ages: young, intermediate, and mature forests. A total of 21,315 data samples are selected, where
training (85\%) and testing (15\%) locations are spatially separated, with held-out grid cells used for evaluation.

\begin{figure}
    \centering
    \includegraphics[width=0.7\linewidth]{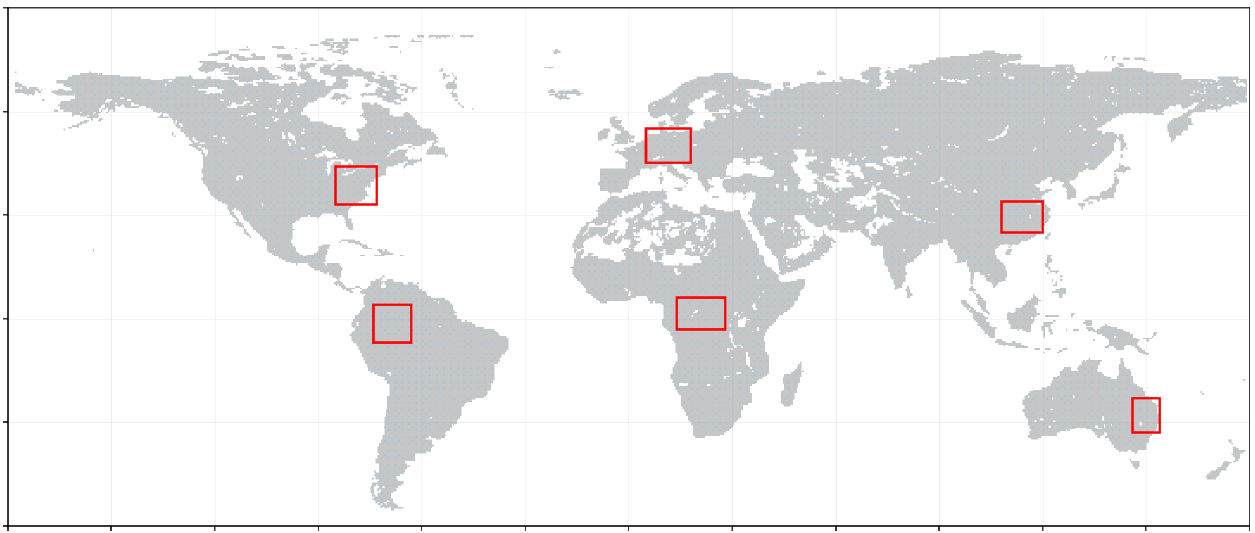}
    \vspace{-10pt}
    \caption{Study areas over six continents.}
    \label{fig:studyarea}
    \vspace{-15pt}
\end{figure}

\textbf{Model configuration.}
At each step, the model takes a five-year history of environmental forcings, which are first embedded by an encoder (MLP) and combined with ecosystem states. A recurrent network (using a standard GRU as an example in this prototype study) then encodes them into latent temporal context, followed by a natural-dynamics head (MLP) to predict height changes, and a decoder (MLP) that predicts the remaining multi-variable responses conditioned on that updated height and latent context.

\textbf{Baselines and metrics.}
We compare against a persistence baseline, $\widehat{\mathbf{y}}_{t+1}=\mathbf{y}_{t}$, and a dedicated no-action baseline emulator trained only for prescribed prediction.
We report long-horizon autoregressive rollout performance using RMSE and relative RMSE (rRMSE), with the latter normalized by variable means to account for scale differences.
For action-conditioned evaluation, we perturb the transition-derived height action to assess whether the model responds sensitively to structural actions and propagates coherent changes to non-height ecosystem variables.

\subsection{Results}

\textbf{Prescribed Emulation Performance.}
We first evaluate whether action-conditioned world model pretraining preserves baseline prescribed emulation performance when no user action is provided. Table~\ref{tab:rollout_rrmse} reports long-horizon rollout rRMSE across young, intermediate, and mature forest conditions.
The World Model achieves performance comparable to, and in several cases better than, the dedicated baseline emulator, while consistently outperforming the persistence baseline.
This indicates that introducing action-conditioned training does not compromise standard prescribed emulation and may provide useful transition-level regularization.

\begin{table}
    \centering
\caption{Rollout mean rRMSE across 3 forest conditions.}
\vspace{-10pt}
\label{tab:rollout_rrmse}
\resizebox{\columnwidth}{!}{
\begin{tabular}{lcc}
\toprule
Model & All Time Steps & Final Time Step \\
& \small (Young / Inter. / Mature) & \small (Young / Inter. / Mature) \\
\midrule
Persistence & 0.710 / 0.295 / 0.302 & 1.124 / 0.376 / 0.347 \\
Baseline emulator & 0.180 / 0.211 / 0.216 & 0.207 / 0.217 / 0.242 \\
World Model & 0.162 / 0.189 / 0.221 & 0.202 / 0.205 / 0.246 \\

\bottomrule
\end{tabular}
}
\vspace{-5pt}
\end{table}

\begin{figure}[h]
    \centering
    \includegraphics[width=1\linewidth]{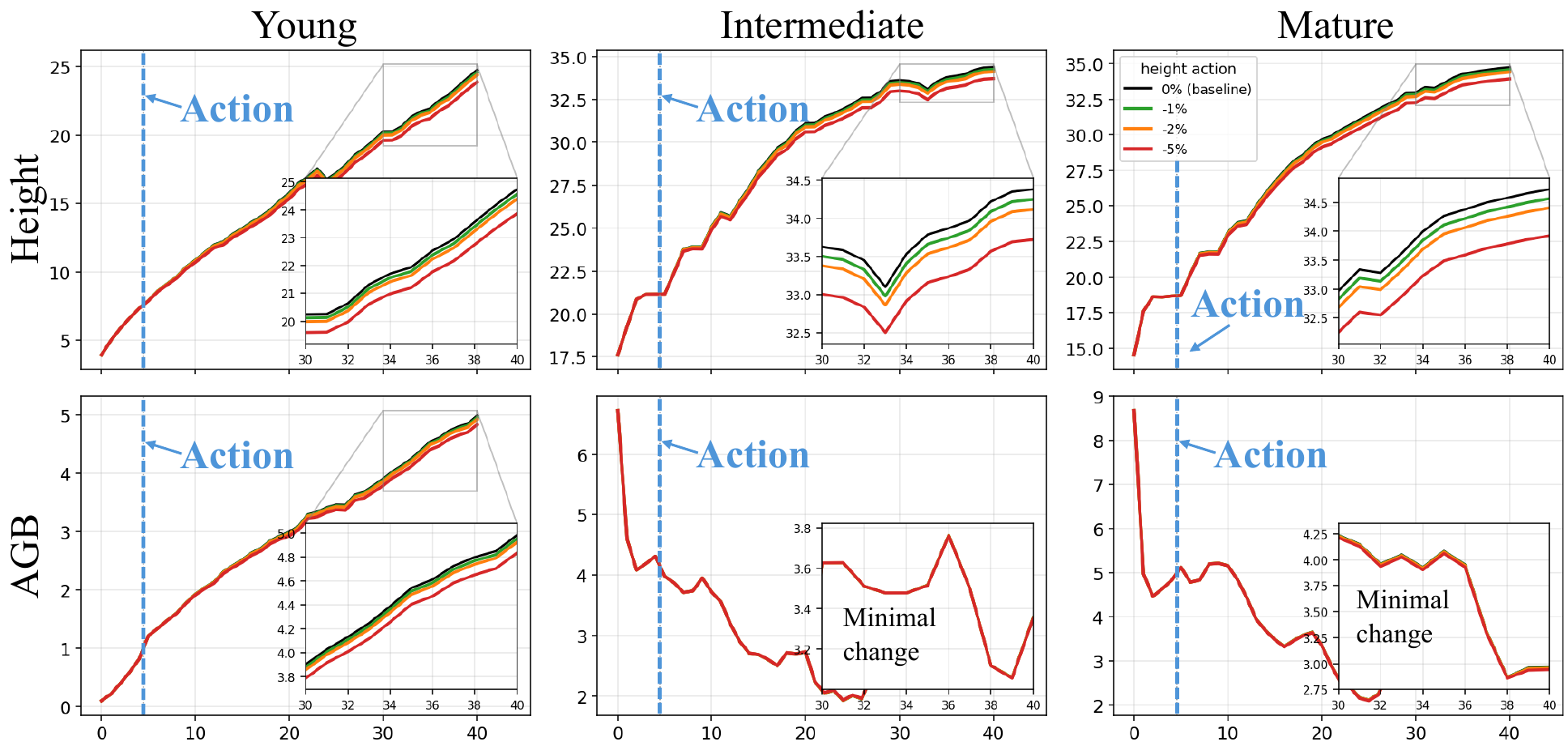}
    \caption{Action-conditioned responses under perturbed height actions (top: height; bottom: aboveground biomass).}
    \label{fig:action}
    \vspace{-10pt}
\end{figure}

\textbf{Controllability under Height Actions.}
We evaluate controllability by perturbing the transition-derived height action and rolling out the model under the modified action.
As shown in the top row of Fig.~\ref{fig:action},
the model produces ordered height responses across perturbation magnitudes, indicating sensitivity to user-specified structural actions.
The effect is stronger in young forests and weaker in mature forests, consistent with age-dependent forest growth dynamics and slower structural change near maturity.

\textbf{Masked Response Inference.}
We further evaluate whether the model can infer coupled ecosystem responses when only the controlled height variable is observed at the action step.
As shown in the bottom row of Fig.~\ref{fig:action},
the deviations of a non-height variable, aboveground biomass, remain close to zero after masking, indicating that the masked model produces responses consistent with the full-observation setting.
In mature forests, non-height responses remain close to zero despite height perturbations, suggesting weaker cross-variable sensitivity.
This suggests the model learns cross-variable dependencies and can infer coupled ecosystem responses from partial state information rather than relying on direct observation of all variables.

\section{Conclusions}

We introduce an action-conditioned world-modeling framework for Earth-system emulation, enabling learned emulators to support both prescribed prediction and user-specified structural interventions.
The proposed framework derives action supervision from state transitions and uses masked response learning to infer coupled ecosystem responses from partial state information.
Experiments on ecosystem trajectories show that this design preserves strong long-horizon emulation performance while enabling consistent responses to user-specified actions and reliable reconstruction of non-controlled variables under masking.
These findings provide a first step toward interactive, intervention-aware Earth-system emulators for scientific what-if analysis.

\begin{acks}
Zhihao Wang, Ruichen Wang, Ruohan Li, and Yiqun Xie are supported in part by
NSF Grant No. 2126474, 2147195, 2425844, and 2530610;
NASA Grant No. 80NSSC25K0013 and 80NSSC25K7221;
Google’s AI for Social Good Impact Scholars program;
and the Zaratan cluster at the University of Maryland.
Xiaowei Jia is supported in part by
NSF Grant No. 2239175, 2147195, 2316305, 2425845, 2530609, and 2203581;
NASA Grant No. 80NSSC24K1061 and 80NSSC

\noindent25K0013;
USGS Grant No. G21AC10564 and G22AC00266;
and Pitt Momentum Funds and CRC at the University of Pittsburgh.
George Hurtt is supported by NASA Grant No. 80NSSC25K7221 and 80NSSC22K1733.
Lei Ma is supported by NASA Grant No. 80NSSC25K7221, 80NSSC24K0599, and 80NSSC24K1632, and Schmidt Sciences.
Gengchen Mai is supported by NSF Grant No. 2521631.
Shaowen Wang is supported by NSF Grant No. 2118329.
We would also like to thank the Derecho system from NSF NCAR.

\end{acks}

\bibliographystyle{ACM-Reference-Format}
\bibliography{references}

\end{document}